\documentclass[conference]{IEEEtran}
\IEEEoverridecommandlockouts
\usepackage{cite}
\usepackage{amsmath,amssymb,amsfonts}
\usepackage{graphicx}
\usepackage{textcomp}
\usepackage{xcolor}
\usepackage[english]{babel} 
\usepackage{graphicx}
\usepackage{subcaption}
\usepackage{graphicx} %package to manage images
\graphicspath{ {./ITSC_FIG/} }
\usepackage[rightcaption]{sidecap}

\usepackage{graphics} % for pdf, bitmapped graphics files
\usepackage{epsfig} % for postscript graphics files
\usepackage{mathptmx} % assumes new font selection scheme installed
\usepackage{times} % assumes new font selection scheme installed
\usepackage{amsmath} % assumes amsmath package installed
\usepackage{amssymb}  % assumes amsmath package installed
\usepackage[utf8]{inputenc}
\usepackage{mathrsfs}
\usepackage{amssymb }
\usepackage{amsfonts}
\usepackage{mathrsfs} 
\usepackage{comment}
\usepackage{titlesec}
\usepackage{amsmath,amsfonts,amssymb}
\usepackage{algorithm2e}
\usepackage[noend]{algpseudocode}
\usepackage{multirow}
\usepackage{longtable}
\usepackage{hyperref}
\usepackage{caption}
\def\BibTeX{{\rm B\kern-.05em{\sc i\kern-.025em b}\kern-.08em
    T\kern-.1667em\lower.7ex\hbox{E}\kern-.125emX}}

\DeclareMathOperator*{\minimize}{minimize} 
\makeatletter
\newcommand\subsubsubsection{\@startsection{paragraph}{4}{\z@}{-1ex\@plus -0.5ex \@minus -.25ex}{1.ex \@plus .25ex}{\normalfont\normalsize\bfseries}}
\newcommand\subsubsubsubsection{\@startsection{subparagraph}{5}{\z@}{-1ex\@plus -0.5ex \@minus -.25ex}{1.ex \@plus .25ex}{\normalfont\normalsize\bfseries}}
\newcommand{\RomanNumeralCaps}[1]
    {\MakeUppercase{\romannumeral #1}}
\makeatother    

\begin{document}
\title { \vspace {15 pt} \LARGE GTP-UDrive: Unified Game-Theoretic Trajectory Planner and Decision-Maker for Autonomous Driving in Mixed Traffic Environments\\}

% \and
% \and

\author{Nouhed Naidja$^{1,2}$, Marc Revilloud$^{3}$, Stéphane Font$^{2}$,  Guillaume Sandou$^{2}$% <-this % stops a space
\thanks{$^{1}$ {Institut VEDECOM, (nouhed.naidja@vedecom.fr)}}
\thanks{$^{2}$ CentraleSupelec, Laboratoire des signaux et systemes (L2S),
(nihed.naidja@centralesupelec.fr), (stephane.font@centralesupelec.fr), 
 (guillaume.sandou@centralesupelec.fr)}\thanks{$^{3}$ {Dotflow, (marc.revilloud@dotflow.fr)}} 
\thanks{*The authors acknowledge the infrastructure and the support of the SCARLET team of Vedecom institute.
Special thanks to Benoît Lusetti and Alexis Warsemann for their involvement in the implementation and experimentation. }
}

\maketitle

% \begin{abstract}
% Understanding the mutual dependence between autonomous vehicles and human drivers is still an open problem with direct implications for the safety and feasibility of autonomous driving.
% This dependence is due to the interactions between traffic participants. Thus, it is crucial for Autonomous Vehicles (AVs) to understand and analyze human-driven vehicles’ intentions and to have a behavior comprehensible to other road users.
% To this end, this paper presents a unified game-theoretic trajectory planner and decision-maker considering a mixed-traffic environment. Our model considers other vehicles’ intentions in the decision-making process, and provides the AV with a human-like trajectory, based on the clothoid interpolation technique. This study uses a solver based on Particle Swarm Optimization (PSO) that fast converges to an optimal decision. 
% Among the highly interactive traffic scenarios, the intersection crossing is particularly challenging. Thus, we choose to demonstrate the feasibility and effectiveness of our method at unsignalized intersection scenarios.
% Testing results show that our approach is suitable for: 1) making decisions and generating trajectories simultaneously. 2) Describing the trajectory as a piecewise clothoïd and enforcing geometric constraints on path curvature. 3) Reducing the dimension of search space for the trajectory optimization problem.
% \end{abstract}

\begin{abstract}
Understanding the interdependence between autonomous and human-operated vehicles remains an ongoing challenge, with significant implications for the safety and feasibility of autonomous driving.
This interdependence arises from inherent interactions among road users.
Thus, it is crucial for Autonomous Vehicles (AVs) to understand and analyze the intentions of human-driven vehicles, and to display behavior comprehensible to other traffic participants.
To this end, this paper presents GTP-UDRIVE, a unified game-theoretic trajectory planner and decision-maker considering a mixed-traffic environment. Our model considers the intentions of other vehicles in the decision-making process and provides the AV with a human-like trajectory, based on the clothoid interpolation technique.
% This study investigates a solver based on Particle Swarm Optimization (PSO) that quickly converges to an optimal decision.
Among highly interactive traffic scenarios, the intersection crossing is particularly challenging. Hence, we choose to demonstrate the feasibility and effectiveness of our method in real traffic conditions, using an experimental autonomous vehicle at an unsignalized intersection. Testing results reveal that our approach is suitable for 1) Making decisions and generating trajectories simultaneously. 2) Describing the vehicle’s trajectory as a piecewise clothoid and enforcing geometric constraints. 3) Reducing search space dimensionality for the trajectory optimization problem.

\end{abstract}

% \begin{IEEEkeywords}
% Autonomous Driving, Decision-Making, Trajectory Optimization,  Particle Swarm Optimization, Game Theory.
% \end{IEEEkeywords}

%%%%%%%%%%%%%%%%%%%%%%%%%%%%%%%%%%%%%%%%%%%%%%%%%%%%%%%%%%%%%%%%%%%%%%%%%%%%%%%%

\section{INTRODUCTION}
Autonomous Vehicles (AVs) are expected to gradually join the transportation realm, coexisting with conventional vehicles throughout a transition period \cite{khan2022integrating}.
Although this cohabitation has great potential, it is still not straightforward and may lead to uncertain interactions\cite{olaverri2020promoting}.
First, given the complexity and the unpredictability of human behavior on the road and second, due to the possibility of misinterpreting autonomous vehicles' decisions by other traffic participants. 
Therefore, it becomes imperative for AVs to understand and account for interactions with human drivers, while also exhibiting a behavior recognizable by other road users \cite{mahajan2021intention}.
In light of this context, the research questions that motivate our work are:
Would emulating a human-like decision process assist AVs in understanding and accounting for human drivers' intentions and decisions? Which behavior should autonomous vehicles exhibit to garner human drivers' acceptance? 

Within the scope of this study, we focus on instances where autonomous vehicles display behavior that is both comprehensible to humans and conforms to road traffic rules.

In this paper, we first adopt a human-like clothoid-based model to describe the trajectories of turning vehicles. Second, we emphasize the utilization of game theory to incorporate the social dynamics embedded in strategic interactions between AVs and human drivers.
The main contribution of our work is a novel approach where the decision-making and trajectory generation are merged in a cohesive framework. The distinctive aspect of our model is that the decision-maker assesses the feasibility of trajectories, and the generated trajectory triggers the update of the decision-making process.

This paper is an extension of our former work presented in \cite{naidja2023interactive}.
It is structured as follows: section \RomanNumeralCaps{1} and section \RomanNumeralCaps{2} provide an introduction and a literature overview, respectively. In section \RomanNumeralCaps{3}, we expose the proposed trajectory generation methodology, followed by a comprehensive description of the decision-making process: starting from the game formulation to the players' payoff design. Section \RomanNumeralCaps{4} showcases the experimental results and the validation of the proposed approach through both implementation of our algorithm using a prototype autonomous vehicle under real road scene conditions, and a Matlab simulation of an unsignalized intersection crossing.
Finally, section \RomanNumeralCaps{5} concludes the paper.
\section{RELATED WORK}
% In the section below, we present a succinct state-of-the-art analysis regarding trajectory planning and decision-making for autonomous vehicles in urban traffic scenarios.
\subsection {\textbf{Vehicle interaction and decision making}}
Autonomous cars should be aware of human drivers' intentions and act consistently with respect to their expectations.
% Latest studies rely on game theory based approaches, and data-driven artificial intelligence (AI) methods, to address decision-making in autonomous driving systems \cite{valiente2022robustness}. Game theory focuses on strategic interactions between agents to find equilibrium solutions, while data-driven AI techniques leverage large datasets to learn optimal policies, enabling adaptive and context-aware behavior.
% In this context, the authors in \cite{10101707} demonstrated throughout comparative studies, that game-theoretic based frameworks have better robustness for strategic interactions than reinforcement learning (RL), and control barrier function CBF) approaches when surrounding vehicles are not safety-conscious. 
Game theory provides rigorous mathematical frameworks to analyze and understand strategic interactions among multiple decision-makers, where the outcomes depend on the choices made by all parties \cite{li2022review}.
By considering the interactions and interdependencies among different road users, game theory aids to predict the outcomes of decisions made by these agents \cite{zhang2010review}.
In this context, the two-player game-based model formulated in \cite{sadigh2016planning} enables efficient coordination between AVs and human drivers. However, the study was limited to the "going straight" scenario, and turning movements were not considered.
More promising in this regard, the paper \cite{tian2018adaptive} considers a complete vehicle model and includes kinematics constraints while designing a decision-making algorithm for autonomous roundabout passing.
% Recently, by extending the game theory approach, cognitive hierarchy reasoning, also called "level-k game", is explored in \cite{kawagoe2012level} to handle the interactive vehicle behavior.
% In this work, the coordination of the vehicles passing the intersection was well achieved, but the authors did not consider the optimization of passing cars' trajectories.
Finally, the study \cite{cleac2022algames} from Stanford University, introduced a trajectory optimization problem in a Nash-style, dynamic game. The algorithm outperforms benchmark algorithms in complex driving scenarios.
The proposed game solver guaranteed local convergence, but not a global one. 

Given the aforementioned studies, we believe that game-theoretic approaches provide a suitable framework for designing a decision-making system that accounts for interactions with human drivers. 
Our contribution regarding the decision-making module builds upon Stanford University's study \cite{cleac2022algames}. 
\subsection{\textbf {Modeling of vehicle turning maneuver}}

Autonomous vehicle's turning maneuver has given rise to significant research work \cite{chouhan2018autonomous}.
The study in \cite{oh2022sharable} highlights that spline interpolation, Bézier curves, and clothoid-based models are particularly suitable for vehicle navigation in complex road geometries. 
% These techniques aim to approximate curves by smoothly connecting a given set of waypoints \cite{gul2021consolidated}. 
% In this context, the paper \cite{lian2020cubic} presents a cubic spline interpolation-based approach for collision-free path planning.
% The authors verified the feasibility of the optimal planned path, but curvature constraints were not considered. Thus, discontinuities could arise at the junction points of the generated path, inducing sudden jerks and impacting the driving comfort \cite{7274361}.
% In the same vein, \cite{yoon2018spline} model the required turning space of car-like vehicles using cubic Bezier curves and a shape-aware spline-based Rapidly-exploring Random Tree \(RRT^{*}\) algorithm concerning the nonholonomic constraints. In these two studies, the proposed algorithms mainly focus on generating collision-free paths. Curvature constraints were not considered. Thus, discontinuities could arise at the junction points of the generated path, inducing sudden jerks and impacting the driving comfort \cite{7274361}.
Clothoid-based interpolation provides a linear, continuous, and smooth curvature variation along the curve. 
This characteristic offers better comfort and maneuverability, prevents undesirable jerks, and allows smooth curvature transitions while entering or exiting a curved road section from a straight road segment\cite{lambert2019optimal}.
Furthermore, clothoid curves are already used by road network design standards \cite{transportation2011policy}. 
Lastly, the study in \cite{gim2016parametric} highlights that clothoid-based models provide accurate representations of human drivers' turning maneuvers, and can handle non-holonomic and dynamic constraints. 
Consequently, AVs utilizing clothoid-based interpolation can interact more effectively with human drivers, reducing the potential for misunderstandings or unpredictable behavior. 
The reasons above motivate numerous researchers to investigate clothoids interpolation for AVs navigation. For instance, the solution presented in \cite{kim2017modified} minimizes both route length and curvature jumps by connecting two clothoids.
This approach uses a modified Q-learning algorithm to approximate the waypoint linking the two clothoids.
Also, the authors in \cite{abdeljaber2020extraction} used a Convolutional Neural Networks-based tool that extracts left-turning trajectories from real traffic recordings. 
The extracted trajectories consist of straight lines, circular arcs, and clothoids.
Choosing among interpolation techniques depends on the requirements of the road scenarios, and the desired driving behavior.
In light of our review of the existing literature, we posit that clothoid-based interpolation best meets the specific requirements and objectives of our study.

\section{GTP-UDRIVE: Unified Trajectory Planner and Decision-Maker for Autonomous  Driving}
Autonomous vehicles display their intentions and actions to human drivers through their trajectories \cite{wang2021decision}.
These trajectories stem from the choices taken during the decision-making process. This tight connection between decision-making and trajectory generation requires a well-coordinated relationship to ensure the systems' reliability.
Consequently, we developed GTP-UDRIVE, a unified decision-making and trajectory planning framework. In this framework, a safety and efficiency-aware decision-maker considers the feasibility of the trajectory. The performed trajectory, in turn, triggers the update of the decision-making process. 

In this section, we present a comprehensive overview of our contributions. The first subsection focuses on the trajectory generation methodology, discussing the technique used to generate optimal trajectories.
The second subsection will delve into the decision-making algorithm, outlining its consideration of human-operated vehicle's intentions in the decision-making process. 
\begin{figure}
\begin{center}
\includegraphics[width=\linewidth]{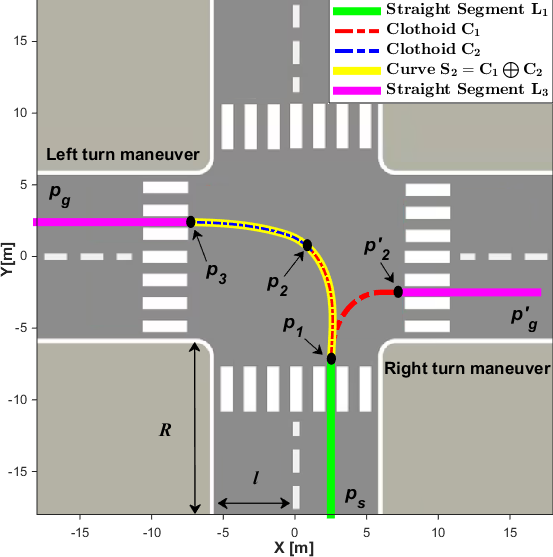}
\caption{Trajectory description: Starting from an initial position \(\textbf{p}_{s} (x_{s},y_{s},\theta_{s})\) and aiming to reach the desired destination  \(\textbf{p}_{g} (x_{g},y_{g},\theta_{g})\), we investigate the conflict zone space to generate feasible trajectories} 
\vspace*{-8mm}
\label{fig:Trajectoire}
\end{center}
\end{figure}

\subsection {\textbf {Clothoid Fitting for Trajectory Generation}} \label{subsection}
 Vehicle trajectories are often represented by a multitude of variables or parameters. Exploring the navigable space can lead to many possible trajectories, resulting in a significant number of optimization parameters. However, trajectory optimization within this expansive search space is computationally intensive and time-consuming. 
To address the complexity of the optimization problem, we propose to reduce the dimensionality of the search space.
To achieve this, instead of explicitly optimizing every point along each possible trajectory, we choose to represent the trajectories by a set of waypoints \(\left\{ {\mathit{p}_{1}},{\mathit{p}_{2}},{\mathit{p}_{3}} \right\} \in \mathbb{R}^{3^3}, {\mathit{p}_{i} }\left ( {x_{i}},{y_{i}},{\theta_{i}} \right ) \in \mathbb{R}^{3}\), \(i=\left\{1,2,3\right\}\) as illustrated in figure \eqref{fig:Trajectoire}.
This shifts the optimization focus toward determining the optimal positions of the waypoints. 
After that, we use a clothoid-fitting method to generate a
smooth trajectory passing through the optimized waypoints.
For further details on the clothoid construction procedure, we
refer readers to our previous work \cite{naidja2023interactive}.

We derived inspiration from \cite{abdeljaber2020extraction} to describe a trajectory \(\textbf{T}\) as a succession of a straight segment \(\mathscr{L}_{1}\) followed by a curve segment \(\mathit{S}_{2}\), and ending by another straight segment \(\mathscr{L}_{3}\), we note: \begin{math} {T} = \mathscr{L}_{1}\bigoplus \mathit{S}_{2} \bigoplus \mathscr{L}_{3}\end{math}.

In the case of a \textbf {right} turn maneuver, the curved segment consists of an elementary clothoid \(\mathscr{C}_{2}=\mathit{C_{1}} \). 
Whereas, for a \textbf {left} turn maneuver, the curve segment \(\mathit{S}_{2}\) is formed by combining two elementary clothoids \begin{math} \mathit{S}_{2} =\mathit{C_{1}} \bigoplus \mathit{C_{2}} \end{math} (dashed blue and red segments illustrated in figure \eqref{fig:Trajectoire}).

We choose to address the \textbf {left} turn maneuver due to its complexity.
 Our solution is adaptable to the right turning maneuver as well. 

Discontinuities may arise at the bridging waypoints due to the difference in curvature between the different segments.
To tackle this problem, we ensure that the position \(({x},{y})\), the tangent direction \(\theta \), and the curvature \(\kappa \) at the \(i^{th}\) segment's end-point, and at the starting point of the \((i+1)^{th}\) segment, are identical by holding the following \cite{bertolazzi2018g2}: 
\begin{align*}&\begin{matrix}\forall {\mathit{p}_{i}} ,\\\end{matrix}\left\{\begin{matrix}&x_{i+1}(0)= x_{\mathit{{i}}}(\mathit{L_{{\mathscr{C}_{i}}}})  \\&y_{i+1}(0)= y_{i}(\mathit{L_{{\mathscr{C}_{i}}}})  \\&\theta_{i+1}(0)= \theta_{i}(\mathit{L_{{\mathscr{C}_{i}}}})  \\ &\kappa_{i+1}(0)= \kappa_{i} (\mathit{L_{{\mathscr{C}_{i}}}})  &\end{matrix}\right.\tag{1} \label{1}\end{align*} 

Where \((\mathit{L_{{\mathscr{C}_{i}}}}) \) denotes the length of the segment \(\mathscr{C}_{i}\).

Also We ensure the uniqueness of the curve \begin{math} \mathit{S}_{2} \end{math} by generating clothoid segments that satisfy the system \eqref{2}, following \cite{bertolazzi2015g1}:
\begin{align*}&\left\{\begin{matrix}x_{\mathit{S}_{2}}(0)= x_{1},& x_{\mathit{S}_{2}}(L_{\mathit{S}_{2}})= x_{3} \\
y_{\mathit{S}_{2}}(0)= y_{1},& y_{\mathit{S}_{2}}(L_{\mathit{S}_{2}})= y_{3} \\ 
\theta_{\mathit{S}_{2}}(0)= \theta_{1},& \theta_{\mathit{S}_{2}}(L_{\mathit{S}_{2}})= \theta_{3} \\
\kappa_{\mathit{S}_{2}}(0)= \kappa_{1},& \kappa_{\mathit{S}_{2}}(L_{\mathit{S}_{2}})= \kappa_{3} &\end{matrix}\right.\tag{2} \label{2}\end{align*}

In addition, we constrain the point where the vehicle starts its turning maneuver at  \({p}_{1} =\left ( \frac{\mathit{l}}{2}, \mathit{R},\frac{\pi}{2}\right ) \), where \(\mathit{l}\) is the lane width, and \(\mathit{R}\) is a characteristic distance before entering the intersection (see figure \eqref{fig:Trajectoire}).   
Finally, we assume that the vehicles' destination is predetermined. Consequently, the tangent angles \(\theta_{2}\) and \(\theta_{3}\) at the locations \(p_{2} \) and, \(p_{3}\), are known. 

Within this structural configuration, four degrees of freedom persist, representing the coordinates of the two remaining waypoints \(p_{2} (\mathit{x}_{2},\mathit{y}_{2})\) and \(p_{3} (\mathit{x}_{3},\mathit{y}_{3}) \). 
These coordinates form a reduced set of new waypoints denoted as: \({\mathbb{W}_{p}} =\left\{(\mathit{x}_{2},\mathit{y}_{2}),(\mathit{x}_{3},\mathit{y}_{3}) \right\} \in \mathbb{R}^{4}\).
Identifying this set is both necessary and sufficient to fully define the trajectory \({T}\). 

In the next section, we introduce a decision-making algorithm that explores a four-dimensional search space, to find the optimal set of waypoints \({\mathbb{W}^{opt}_{p}}\) described above.

%%%%
%%%%
%%%%

\subsection{\textbf{Decision-Making : A Game-Theoretic Perspective}}
In this paper, we aim to capture social interactions between an AV and a human-driven car. To this end, 
we explicitly incorporate the human driver's decision and intention while designing our decision-making algorithm.
Nash Equilibrium (NE) is a solution concept that captures strategic interactions among multiple players. It identifies a set of strategies, one for each player, in which no player can improve their outcome by independently changing their chosen strategy \cite{holt2004nash}.
In the context of autonomous driving, Nash equilibrium allows autonomous vehicles to make strategic choices that balance their objectives with the actions of other vehicles, leading to smoother traffic and improved safety.
Furthermore, Nash equilibrium models players symmetrically and assumes that all players are rational decision-makers.
Nevertheless, in the case of intersection crossing, drivers' objectives are interconnected by coupled requirements for collision avoidance.
This optimization problem belongs to Generalized Nash Equilibrium Problems (GNEPs), an extension of the original Nash equilibrium.
In GNEPs, the objective functions and constraints of each player are influenced by the strategic space of their opponents \cite{facchinei2010generalized}. A solution of a GNEP is a set of players' strategies satisfying all players' optimization requirements.
The present study proposes a non-cooperative dynamic game between two players. Given the symmetry of Nash-style games, the two players are modeled equivalently throughout this paper.
\vspace{0.18 cm}
\subsubsection{\textbf{Game formulation}}
\vspace{0.12 cm}
 The player \textbf{\textit{V}}, and its opponent \textbf{\textit{O}}, control their strategic spaces 
 \begin{math} \mathcal{S}_{v}\subset \mathbb{R}^{n}\end{math} and \begin{math}
\mathcal{S}_{o}\subset \mathbb{R}^{m}\end{math}, through their \textbf{\textit{n}} and \textbf{\textit{m}} decision variables respectively. 

We define the game joint strategic space as \begin{math} \mathcal{S}^{game}(\mathcal{S}_{v},\mathcal{S}_{o})= \mathcal{S}_{v} \times \mathcal{S}_{o}\in \mathbb{R}^{n\times m} \end{math}, the Cartesian product of pure strategy sets of the two players.

\(\mathit{Q}^{p} ({s}^{v},{s}^{o}): \mathbb{R}^{n\times m}\to \mathbb{R}\) is the payoff function for a player \textbf{\textit{p}}.

Thus, given an opponent strategy \({s}^{o}\), a player \textbf{\textit{V}} aims to solve the optimization problem in equation \eqref{3.a}: 

\begin{equation}
    \begin{aligned} \begin{aligned} \mathop {\minimize }\limits _{{s}^{v} \in \mathcal{S}_{v}} \quad& \mathit{Q}^{v} ({s}^{v},{s}^{o}) \quad& \text {s.t.} \quad& {s}^{v},{s}^{o}\in \mathcal{S}^{game}(\mathcal{S}_{v},\mathcal{S}_{o}) \\ 
    \end{aligned}\end{aligned}
    \tag{3.a}
    \label{3.a}
\end{equation} 
Note that the opponent player's optimization problem is structured similarly.
\begin{equation}
    \begin{aligned} \begin{aligned} \mathop {\minimize }\limits _{{s}^{o} \in \mathcal{S}_{o}} \quad& \mathit{Q}^{o} ({s}^{o},{s}^{v}) \quad& \text {s.t.} \quad& {s}^{v},{s}^{o}\in \mathcal{S}^{game}(\mathcal{S}_{v},\mathcal{S}_{o}) \\ 
    \end{aligned}\end{aligned}
    \tag{3.b}
    \label{3.b}
\end{equation} 

We assume that each player acts rationally and consistently to solve a GNEP.
The rationality we refer to is “the Best Response” concept, commonly accepted in the context of game theory.
It assumes that every player is motivated by choosing the best response available to optimize its own payoff \cite{askari2019behavioral}.
The solution of this GNEP is then an equilibrium \begin{math} ({s}^{*,v},{s}^{*,o})\in \mathcal{S}^{game} \end{math} such that: 
\begin{equation}
\begin{matrix}
\forall  {s}^{v},\forall  {s}^{o}& \left\{\begin{matrix}
\mathit{Q}^{v} ({s}^{v*},{s}^{o*}) \leq \mathit{Q}^{v} ({s}^{v},{s}^{o*})\\\mathit{Q}^{o} ({s}^{v*},{s}^{o*}) \leq \mathit{Q}^{o}  ({s}^{v*},{s}^{o})
\end{matrix}\right.\\
\end{matrix}  
\tag{4}
    \label{4}
\end{equation}
We formulate the generalized Nash equilibrium problem for a non-zero-sum, two-player dynamic game in equation \eqref{5}: 
%%%% proposition 3
\begin{equation}
  \begin{split}
\mathit{J} ({s}^{v},{s}^{o}) & = \max \left\{ \sup_{\overline {{s}^{v}} \in \mathcal{S}_{v}} \left\{\mathit{Q}^{v} ({{s}^{v}},{{s}^{o}})-\mathit{Q}^{v} (\overline{{s}^{v}},{{s}^{o}}) ,0\right\}\right\}+ \\
      & \max \left\{ \sup_{\overline {{s}^{0}} \in \mathcal{S}_{o}} \left\{\mathit{Q}^{o} ({{s}^{v}},{{s}^{o}})-\mathit{Q}^{o} ({{s}^{v}},\overline{{s}^{o}}) ,0\right\}\right\}
\end{split}  
\tag{5}
    \label{5}
\end{equation}

In the first term of the sum, we aim to find the maximum improvement in player V's (the autonomous vehicle) payoff over all possible alternative strategies.
Thus, we analyze the differences between player V's payoff when following its chosen strategy \(\overline {{s}^{v}}\) (i.e. \(\mathit{Q}^{v} (\overline{{s}^{v}},{{s}^{o}}) \)), and \(\mathit{Q}^{v} ({{s}^{v}},{{s}^{o}})\) the payoff when considering a potential deviation to an alternative strategy \({{s}^{v}}\) belonging to \(\mathcal{S}_{v}\), the set of player V's feasible strategies. 

If this difference is positive, it indicates that player \textbf{\textit{V}} could benefit from deviating to an alternative strategy. Conversely, a negative difference indicates that player V's chosen strategy \(\overline {{s}^{v}}\) is optimal, and there is no incentive for deviating.
Player O's payoff is evaluated similarly in the second part of the sum.

Optimizing the problem in equation \eqref{5} enables the identification of the strategic choices \({s}^{v}, {s}^{o}\) leading to the highest payoffs for each player. A generalized Nash equilibrium is attained: both players' objectives are balanced, and no player can improve its payoff by independently changing its strategy.

In our study, each strategy represents a possible set of points of interest \begin{math} {s}_{i}^{v} = \mathbb{W}_{i}^{v}\end{math}, and \begin{math} {s}_{j}^{o} = \mathbb{W}_{j}^{o}\end{math}, for ego player and its opponent respectively.
Thus, the outcomes of a GNEP are two sets of optimal waypoints, one for each player, that ought to be found to generate both AVs and human-driven vehicles trajectories.
This game structure allows for linking the trajectory generation and the decision-making process, thereby building a comprehensive driving unit. 
\vspace{0.18 cm}

\subsubsection{\textbf{Players payoff design}}
\vspace{0.12 cm}
The payoff function \(\mathit{Q}^{p} ({s}^{v},{s}^{o})\) for each player \textbf{\textit{p}} is a linear combination of cost indicators on its trajectory.
We hereby introduce these indicators from the perspective of the ego vehicle. The opponent vehicle's payoff function is derived through symmetry.

\begin{itemize}
\item{\textbf{Efficiency Awareness}}:  %%%%%%%%%%%%%%%%%%%%%%%%%%%%%%%% Efficiency Awareness
To achieve efficiency while crossing the intersection, the AV is prompted to keep a relatively high velocity as outlined in equation \eqref{6} below: 
\begin{equation}
\mathit{I^{eff}_{({s}^{v})}}=\frac{\left|V_{max} - \overline{v}({s}^{v}) \right|}{ V_{max}} \tag{6} \label{6}
\end{equation} \vspace{-0.12 cm}
where \(V_{max}\) is the maximum speed limit permitted within the intersection, and 
\(\overline{v}({s}^{v})\) is the average velocity following strategy \({s}^{v}\). 

\item \textbf{Safety Enhancement}: Continuous Collision Check \newline
We encapsulate each generated trajectory into a convex Oriented Bounding Box (OBB) described by its vertices as \( \left\{\overrightarrow{\mathbf{B_{j}^{v}}} \right\}_{j=1}^{4}= convexhull(\left\{(x_{1}^{v},{y}_{1}^{v}),\cdots,(x_{4}^{v},{y}_{4}^{v}) \right\}) \) and \(\left\{\overrightarrow{\mathbf{B_{j}^{o}}} \right\}_{j=1}^{4}= convexhull(\left\{(x_{1}^{o},{y}_{1}^{o}),\cdots,(x_{4}^{o},{y}_{4}^{o}) \right\})\), representing the ego vehicle and the opponent vehicle bounding boxes, respectively.\newline
We predict potential collisions between the two boxes \( \textbf{B}^{v} \) and \( \textbf{B}^{o} \) by continuously computing the remaining Gap To Collision \((\mathit{\textbf{GTC})}\) at each time \(t\), using an algorithm based on the Separating Axis Theorem (SAT) \cite{han2019local}.
In this algorithm, we identify candidate separating axes, including the normal vectors \(\overrightarrow{N}_{i}\) of the OBBs.
Then, the vertices of both \( \textbf{B}^{v} \) and \( \textbf{B}^{o} \) are projected onto each candidate axis, yielding the minimum and the maximum projections. The projected distance \(D_{proj}\) of a vertex of the ego vehicle's OBB \(\overrightarrow{\mathbf{B_{j}^{v}}}_{j=1}^{4}\) onto a normal vector \(\overrightarrow{N}_{i}\) is given by the expression in equation \eqref{7}:
\begin{multline}
\footnotesize
\begin{matrix}
\hspace{-0.5cm}D_{proj(\overrightarrow{\textbf{B}}_{j}^{v},\overrightarrow{N}_{i})} \\
j\in \left\{ 1,..,4\right\} \\ (i=x,i=y)
\end{matrix} =
\begin{cases}
\left(\left|\frac{\overrightarrow{\textbf{B}}_{j}^{v}\cdot\overrightarrow{\textbf{N}}_{i}}{\overrightarrow{\textbf{N}}_{i}}\right|\right) & \text{if} \quad \frac{\overrightarrow{\textbf{B}}_{j}^{v}\cdot\overrightarrow{\textbf{N}}_{i}}{\left|\overrightarrow{\textbf{N}}_{i}\right|} < 0 \\
\left(\left|\frac{\overrightarrow{\textbf{B}}_{j}^{v}\cdot\overrightarrow{\textbf{N}}_{i}}{\overrightarrow{\textbf{N}}_{i}}\right|-\left|\overrightarrow{\textbf{N}}_{i}\right| \right)& \text{if} \quad \left(\left|\frac{\overrightarrow{\textbf{B}}_{j}^{v}\cdot\overrightarrow{\textbf{N}}_{i}}{\overrightarrow{\textbf{N}}_{i}}\right|-\left|\overrightarrow{\textbf{N}}_{i}\right| \right) > 0 \\
 \quad  \quad   \quad \quad 0 & \quad \quad \text{Otherwise} \\
\end{cases}\tag{7}\label{7}
\end{multline}
where \( j\in \left\{ 1,..,4\right\}\) denotes the index of the vertex, and \( i\) indicates the direction of the normal vector (x or y).
The remaining Gap To Collision \((\mathit{\textbf{GTC})}\) is computed following the equation \eqref{8} as: 
\begin{align}
\hspace{-0.5cm} \mathit{\textbf{GTC}} &= \underset{i}{\max}\left\{ \left( \max\left( D_{\min,i}^{e},D_{\min,i}^{o} \right) - \min\left( D_{\max,i}^{e},D_{\max,i}^{o} \right) \right), 0 \right\} \nonumber \\
\tag{8}\label{8}
\end{align}

With \( D_{min_,i}^{e}\) and \( D_{max_,i}^{e}\) representing the minimum and the maximum extents for the ego vehicle's OBB along the axis \( i\) respectively, and \( D_{min_,i}^{o} \) and \(D_{max_,i}^{o}\) the extents for the opponent vehicle. We utilize this metric to formulae a safety indicator as follows:

\begin{equation}
    \begin{aligned} \mathit{I_{({s}^{v},{s}^{o})}^{safe}} = \begin{aligned} \mathop {\min }\limits _{t_{min}<t<t_{max}} \quad& e^{-\left (\frac{\mathit{GTC(t)}}{\mathit{G_{crit}}} \right )} \\ 
    \end{aligned}\end{aligned}
    \tag{9}
    \label{9}
\end{equation}

\( [t_{min},t_{max}] \) is the duration of interaction between two vehicles. \(\mathit{G_{crit}}\) is a critical threshold that must be maintained consistently between two vehicles.
\end{itemize}

%%%%%

\begin{itemize}
\item {\textbf{Collision Avoidance}:} 

We design an adaptive elliptic safety zone \(\zeta \) that encompasses each vehicle to ensure collision-free navigation. 
At each time, \(t\) we guarantee: 
\end{itemize}
\begin{equation} \forall{t}\in[t_{min},t_{max}], \quad \zeta (t) \bigcap {\textbf{B}}^{o} (t) = \emptyset \tag{10} \label{10} \end{equation} 

% \(\textbf{B}^{o}\) is the oriented bounding box encapsulating the opponent vehicle.   

We continuously check that the constraint in equation \eqref{10}, is respected for each vertex \begin{math} {v}_{j}^{o} ({x}_{j}^o, {y}_{j}^o)\end{math}, belonging to \(\textbf{B}^{o}\), the OBB encapsulating the opponent vehicle by holding the condition in equation \eqref{11} :

% We continuously check that the constraint in equation\eqref{10} for each vertex \begin{math}
%  {v}_{j}^{o} ({x}_{j}^o, {y}_{j}^o)\in \textbf{B}^{o}\end{math},
% \(\forall j\in \left\{1,..,4\right\}\) following:

\begin{equation}
 \begin{aligned} &\frac{\left[ cos(\theta)\cdot ({x}_{j}^{o}-x)+sin(\theta)\cdot ({y}_{j}^{o}-y )\right] ^{2}}{D(s^{v})^{2}}+\\&\frac{\left[sin(\theta)\cdot ({x}_{j}^{o}-x) -cos(\theta)\cdot ({y}_{j}^{o}-y)\right]^{2}}{d^{2}}>  1 \end{aligned}\tag{11} \label{11}   
\end{equation}

Where \((x,y)\) represents the center coordinates of the ellipse \(\zeta (t)\), and \(\theta\) its orientation. These parameters are determined based on the position and orientation of the vehicle's COG.

This elliptic safety zone dynamically evolves according to \(D(s^{v})\) and \(d\), the major and minor semi-axes of the ellipse, respectively. 
\begin{align}
 \normalsize
\left\{\begin{matrix}
D({s}^{v})=\frac{1}{2} {L}_{v}+\text{TTC}\cdot\textit{v}({s}^{v})\\
d=\frac{1}{2} {l}_{v}+d_{safe}\end{matrix}\right.\tag{12} \label{12}
\end{align}

Note that: \(\textit{v}(s^{v})\) is the speed achieved when adopting strategy \(s^{v}\). 
The parameters \(L_{v}\) and \(l_{v}\) correspond to the vehicle's length and wheelbase, respectively. \(d_{safe}\) is a parameter representing a lateral safety distance.
The Time To Collision (TTC), is a safety indicator that quantifies the remaining time until a collision occurs \cite{zhu2020safe}. This metric is employed to adjust the safety zone of the ego vehicle and to ensure that a sufficient distance is maintained from the opponent vehicle. 

\begin{itemize}
    \item \textbf{Respect the lane boundaries}: 
    
The vehicle has to keep driving in the middle of its lane.
Thus, we substitute the vertex  \({v}_{j}^{o}\)'s coordinates \(({x}_{j}^o, {y}_{j}^o)\) in equation\eqref{11} by \((x_{p_{b}},y_{p_{b}})\), the coordinates of the closest point on each road boundary noted \(p_{b}\).
\end{itemize}

We want to minimize the risk of collision and unnecessary stops in the conflict zone.
Thus, we formulated the objective of the player in equation \eqref{13} as a weighted linear combination of safety and efficiency indicators:\begin{equation} f(s^{v},s^{o})=\omega_{1} \cdot \mathit{I_{({s}^{v},{s}^{o})}^{safe}} + \omega_{2}\cdot \mathit{I^{eff}_{({s}^{v})}}, \quad 0 \leq  \omega_{i}\leq 1  \tag{13} \label{13}\end{equation}
This objective is subject to nonlinear constraints associated with the left-turn maneuver as well as the strategies of the opposing vehicle. 
The optimization problem that the player needs to solve can be stated as follows: 
\begin{equation} 
 \begin{aligned} \mathop {\min }\limits _{s^{v},s^{o}} \quad& f(s^{v},s^{o}), & \text {s.t.} \quad & C(s^{v},s^{o}) \leq 0, \\ \end{aligned}  \tag{14} \label{14}\end{equation}
Where \(C(s^{v},s^{o})\) encompasses the inequality constraints outlined in equation \eqref{11}.

We transform the problem in equation \eqref{14} to an unconstrained optimization problem. 
Thus, we add a term penalizing constraints violation, and re-write equation \eqref{3.a} representing the former optimization problem of a player in equation \eqref{15}:
 \begin{equation} \mathit{Q}^{p} ({s}^{v},{s}^{o})= f(s^{v},s^{o})+\mathcal{G}(C(s^{v},s^{o}))\tag{15} \label{15}\end{equation}
 
The penalty function \( \mathcal{G}(C(s^{v},s^{o}))\) is formulated as follows: 
\begin{equation} \mathcal{G}(C)=
\begin{matrix}
\begin{cases}
0 & \text{ if } C(s^{v},s^{o})< 0 \\
\lambda \cdot C(s^{v},s^{o}) & \text{ Otherwise } 
\end{cases}, &   \lambda \in \mathbb{R}^{*+}
\end{matrix}  \tag{16} \label{16}\end{equation}

\( \lambda \) is a penalty coefficient that affects the sensitivity of the optimization algorithm to constraint violations.

Finally, this study investigates the same game solver based on Particle Swarm Optimization (PSO), introduced in our recent paper \cite{naidja2023interactive}, to find a strategic combination (\(s^{v},s^{o}\)) that solves the GNEP in equation \eqref{5}.

\section{EXPERIMENTATION AND SIMULATION RESULTS} 
\begin{figure}[htbp]
    \centering
    \begin{subfigure}[b]{0.45\textwidth}
        \includegraphics[width=\textwidth]{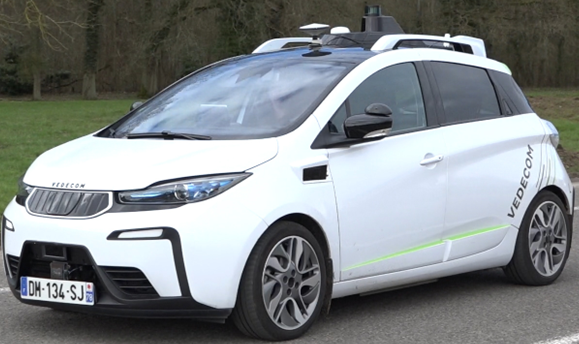} % Include your first figure here
        % \caption{Outside view}
        % \label{fig:sub1}
    \end{subfigure}
    \hfill
    % \begin{subfigure}[b]{0.45\textwidth}
    %     \includegraphics[width=\textwidth]{ITSC_FIG/zoe2.png} % Include your second figure here
    %     \caption{Inside view}
    %     \label{fig:sub2}
    % \end{subfigure}
    \caption{Experimental Autonomous test vehicle of Institut VEDECOM}
    \label{fig:both}
\end{figure}

The experiments have been performed on unsignalized intersection located in All.des Marronniers, Versailles-Satory (78000), France. We validate our trajectory generator using the experimental autonomous vehicle of Institut VEDECOM, which is an electric Renault Zoe equipped with different perception and localization sensors as shown in figure \eqref{fig:both}. 

\subsection{Validation of the Trajectory Generation Methodology}

\textbf{Generated trajectory VS human driver trajectories}

\begin{figure}
    \centering 
    % \leftskip-1em
    \includegraphics[width=\linewidth, height=7.5cm]{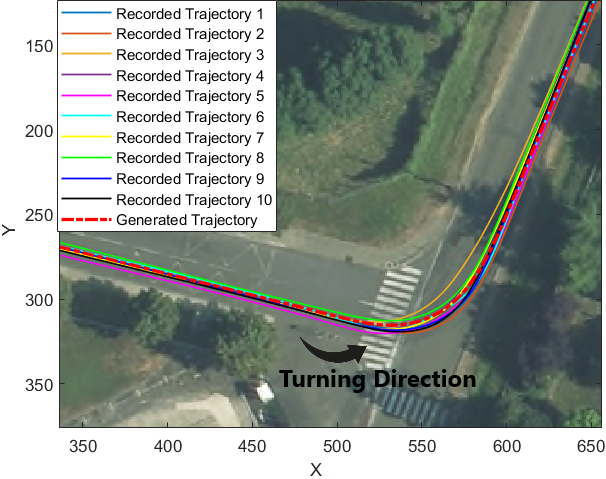}
    \caption{All.des Marronniers intersection: Comparison between recorded left turn human driver trajectories, and GTP-UDRIVE generated trajectory (in dashed red)}
    \vspace*{-5mm}
    \label{fig:humainVsauto}
\end{figure}

To validate whether the proposed trajectory generation methodology provides accurate
representations of human drivers’ turning maneuvers, we recorded 10 left turn human driver trajectories at All.des Marronniers intersection.
The recorded trajectories were performed by a single driver, instructed to drive in the middle of the lane. 
The recordings are compared to a GTP-UDRIVE generated trajectory (represented in dashed red in figure \eqref{fig:humainVsauto}) using identical initial conditions, and geometric layout parameters. These parameters are provided in Table\eqref{tab:table1}.
Comparison results are shown in figure \eqref{fig:humainVsauto}.
% \begin{table}[]
% \centering
% \caption{\label{tab:table1}Simulation Parameters.}
% \begin{tabular}{|cccc|}
% \hline
% \multicolumn{4}{|c|}{\textbf{Simulation Parameters}}              \\ \hline
% \multicolumn{1}{|c|}{Lane width} & \multicolumn{1}{c|}{\({l}= 3.5(m)\)} & \multicolumn{1}{c|}{ Dist to intersection}    & \({R}=80 (m)\)\\ \hline
% \multicolumn{1}{|c|}{Vehicle length} & \multicolumn{1}{c|}{\({L}_{v}=3.9 (m)\)}   & \multicolumn{1}{c|}{Wheelbase}      & \({l}_{v}=1.9 (m)\) \\ \hline
% \multicolumn{1}{|c|}{Time To Collision}  & \multicolumn{1}{c|}{\(TTC =2 (s)\)}   & \multicolumn{1}{c|}{Maximum Speed}  & \({V}_{max}= 13 (m/s)\) \\ \hline
% \multicolumn{1}{|c|}{lateral distance}  & \multicolumn{1}{c|}{\(d_{safe}= 0.2 (m)\)}   & \multicolumn{1}{c|}{penalty coefficient}  & \( \lambda= 10^{3} \) \\ \hline
% \end{tabular}
% \end{table}

The current sample size does not offer a comprehensive validation, yet it can provide an initial validation of the functionality of the proposed method.
\begin{table}[]
\vspace*{-4mm}
\centering
\caption{\label{tab:table1}Simulation Parameters}
\begin{tabular}{|c|c|c|p{1.2cm}|}
\hline
\multicolumn{4}{|c|}{\textbf{Simulation Parameters}}              \\ \hline
\multicolumn{1}{|c|}{ Dist to intersection} & \multicolumn{1}{c|}{\({R}=80 m\)} & \multicolumn{1}{c|}{Lane width} & \({l}= 3.5m\)\\ \hline
\multicolumn{1}{|c|}{Time To Collision} & \multicolumn{1}{c|}{\(TTC =2 s\)}   & \multicolumn{1}{c|}{Wheelbase}      & \({l}_{v}=1.9 m\) \\ \hline
\multicolumn{1}{|c|}{Maximum Speed}  & \multicolumn{1}{c|}{\({V}_{max}=13 m/s\)}   & \multicolumn{1}{c|}{Vehicle length}  & \({L}_{v}=3.9 m\) \\ \hline
\multicolumn{1}{|c|}{Lateral distance}  & \multicolumn{1}{c|}{\(d_{safe}= 0.2 m\)}   & \multicolumn{1}{c|}{Penalty coefficient}  & \( \lambda= 10^{3} \) \\ \hline
\end{tabular}
\end{table}

\subsection{Vehicle Interaction and Decision Making}

\begin{figure}
    \centering
    
    \includegraphics[width=\linewidth, height=7.2cm]{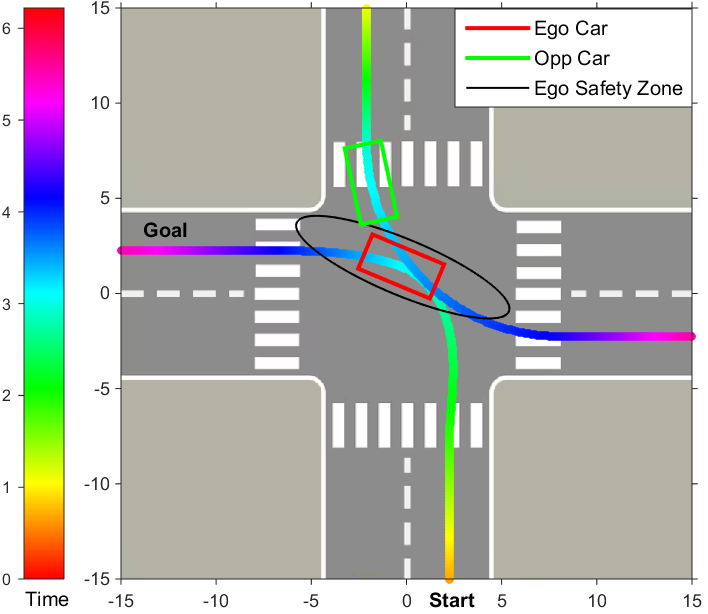} 
    {a) Crossing scenario without running GTP-UDRIVE}
    % \caption{Figure 1}
    
    \vspace{10 pt}
    \includegraphics[width=\linewidth, height=7.2cm]{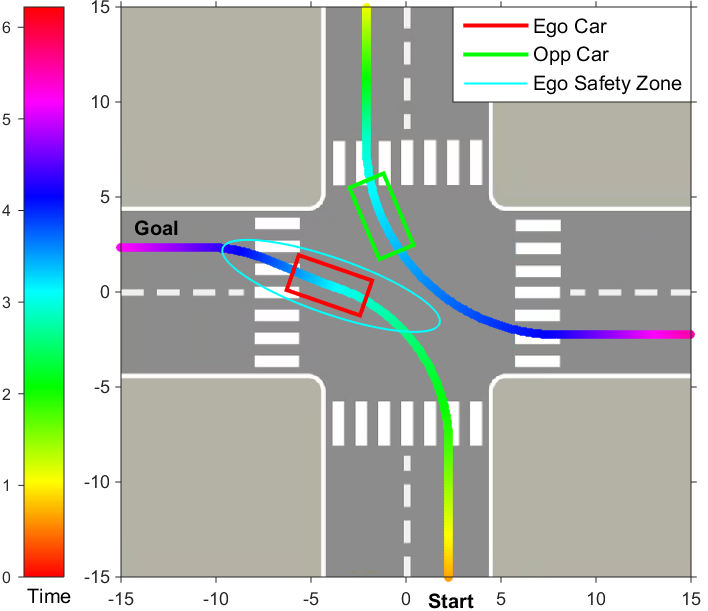} {b) Crossing scenario while running GTP-UDRIVE}
   
    \vspace{6 pt}
    \includegraphics[width=\linewidth, height=7.2cm]{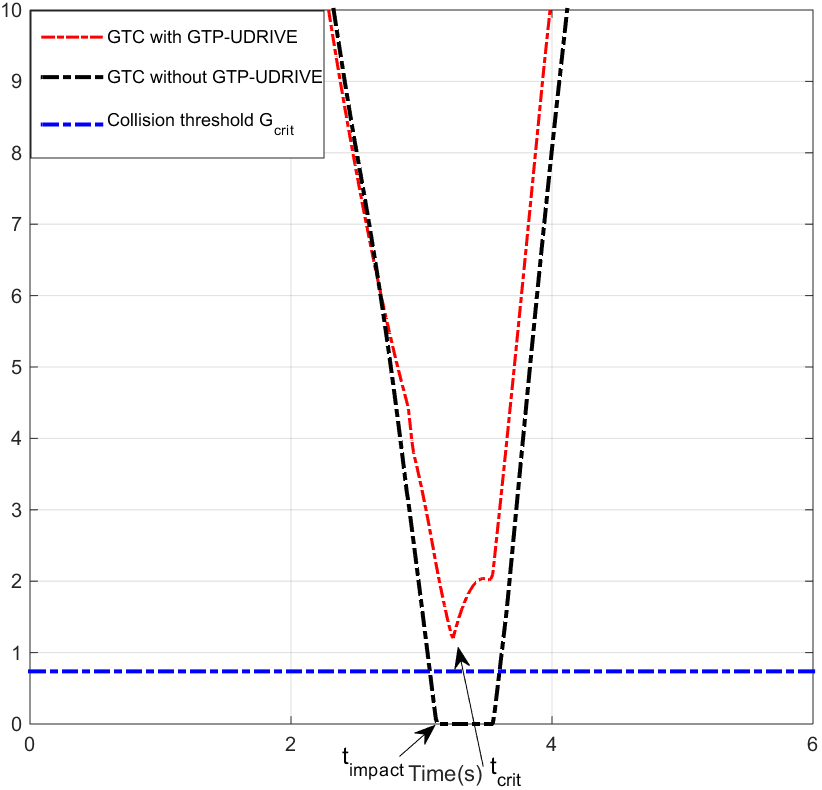} {c) GTP-UDRIVE impact on the Gap To Collide }
    \caption{a) The nominal case at impact time \( \textbf{t}_\textbf{impact}\), b) Post-application of GTP-UDRIVE at \( \textbf{t}_\textbf{crit}\), and c) The evolution of the gap to collision GTC.}
    \label{fig:Before_Optim_Case_1}
\end{figure}

\textbf{Traffic Scenarios} \newline
We consider an unprotected left-turn driving scenario in a mixed traffic environment. This scenario is illustrated in \eqref{fig:Before_Optim_Case_1}, where an autonomous vehicle (in red) is turning left in front of an oncoming human-driven car (in green). 
In our simulation, our main focus lies on scenarios where direct communication between the two vehicles is unavailable.
We exclude Vehicle to Vehicle communication (V2V) as we assume that the human operated vehicle is devoid of V2V communication technology.
Consequently, the autonomous vehicle only relies on perception to estimate its opponent's position and direction.

\textbf{Crossing without running GTP-UDRIVE: Nominal case }

We first simulated the autonomous vehicle and its opponent entering the conflict zone simultaneously. If the AV maintains its initial trajectory, it will lead to a collision with its opponent. Illustration \eqref{fig:Before_Optim_Case_1}.a depicts the collision moment.

\textbf{Crossing while running GTP-UDRIVE}

\begin{figure}
    \centering
      \vspace{6 pt}
    \includegraphics[width=\linewidth, height=7.2cm]{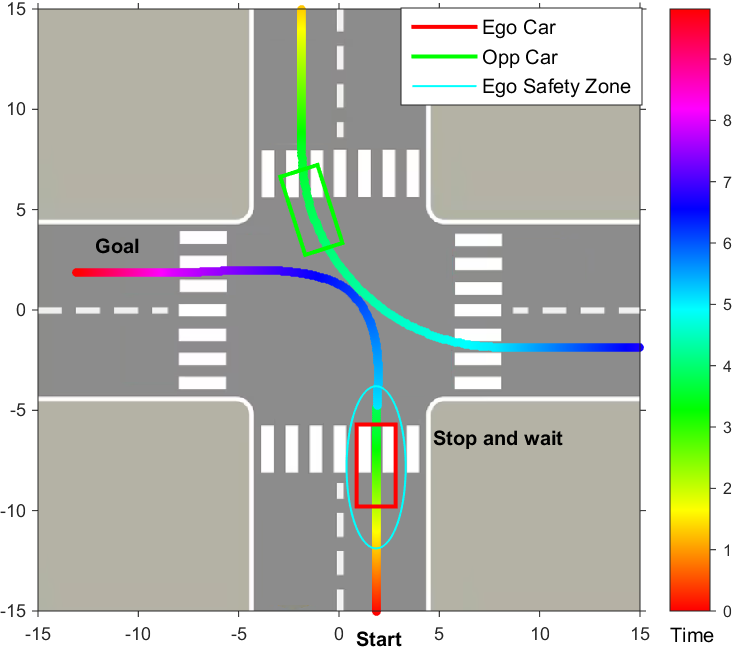} {a) Ego vehicle yield the way resulting from GTP-UDRIVE}
    
    \vspace{6 pt}
    \includegraphics[width=\linewidth, height=7.2cm]{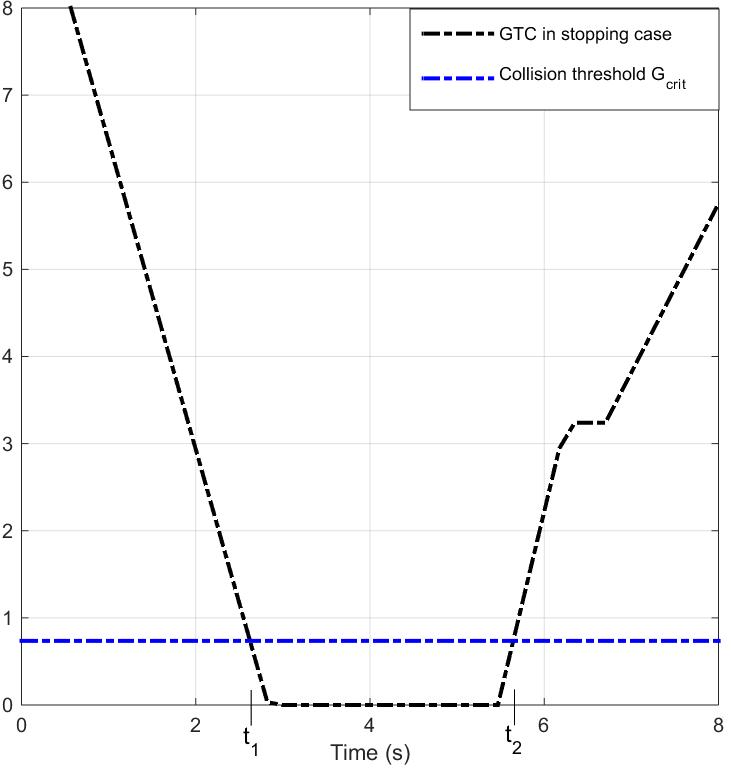} {b) Evolution of the Gap To Collide }
    
    \caption{Case 2: Ego vehicle stops between \({{t}_{1}}= 2.85 (s)\) and \({{t}_{2}}= 5.76 (s)\) until the opponent vehicle has safely cleared the conflict zone.} \vspace*{-3mm}
      \label{fig:Gap_case_2}
\end{figure}

We run our algorithm GTP-UDRIVE, preserving the nominal scenario's simulation parameters and initial conditions.
The AV uses these informations to generate the trajectory of the opponent vehicle following its own trajectory generation methodology described in the section \eqref{subsection}. 
Leveraging the Nash equilibrium symmetry, the AV estimates the opponent's car cost using the objective function in equation \eqref{15} and thus, solves the GNEP in equation \eqref{5}. 
The observed behaviors are illustrated in the figures \eqref{fig:Before_Optim_Case_1} and \eqref{fig:Gap_case_2}, and detailed below: 

\textbf{Case 1}: 
The figure \eqref{fig:Before_Optim_Case_1}.b illustrates the first behavior. The two vehicles decided to cross the intersection without stopping, since they predicted that the two of them would choose a free-conflict trajectory. 
For the ego car (in red), the solution reveals optimal junction points \(\mathbb{W}^{opt}_{p}=\left\{ {\mathit{p}^{*}_{1}},{\mathit{p}^{*}_{2}}\right\}\), enabling the adjustment of the initial trajectory to ensure a collision-free path. 
We observe that the gap to the collision at the critical instant \(\textbf{t}_{\textbf{crit}}\), when the vehicles are closest, ensures that the AV remains outside the collision zone (see figure \eqref{fig:Before_Optim_Case_1}.c).
% Consequently, The ego vehicle decides to cross the intersection first.

\textbf{Case 2}: 
This time, the opponent vehicle (the green, human-operated car) enters the conflict zone prior as in figure \eqref{fig:Gap_case_2}.a.
In this situation, the ego vehicle does not found an alternative trajectory that would enable it to avoid conflict with the trajectory to be executed by the opposing vehicle (Opp). The gap to collision is lower than the minimum safety threshold (see figure \eqref{fig:Gap_case_2}.b), ego vehicles' best response is to stop during the time interval \([t_{1},t_{2}]\), and wait until the opponent vehicle has safely cleared the conflict zone.

\section{CONCLUSIONS}

This paper presents GTP-UDRIVE, a unified decision-maker and trajectory optimizer framework based on the game theory paradigm. 
Our main objective was to foster the coexistence of autonomous and human-driven vehicles in a mixed environment and to enhance road safety and efficiency.

We aimed to provide autonomous vehicles with an efficient decision-making framework that allows the consideration of human drivers decisions.

We demonstrate that our algorithm is effective in planning for an autonomous vehicle to negotiate complex driving scenarios while interacting with other vehicles.
The main advantage of our method is combining the search space for trajectory optimization with the strategic space of the vehicles.
The results from the proposed model validation are promising, as predicting other participants' intentions allows the two vehicles to cross the intersection safely and avoid an unnecessary stop.

From these results, it is reasonable to conclude that the unified framework for trajectory optimization and decision-making proposed in our work is appropriate to address intersection crossing. 

\textbf{Limitations and future work} \newline
Future work will expand the obtained results to consider the interaction with connected vehicles and other autonomous vehicles in the road scene. More attention will be given to experimental validations. 
It would be interesting to include the presented results in an adaptive control-type approach and to consider the evolution of the accurate trajectory of other drivers.
Lastly, one could envision exploring different notions of rationality and considering other behaviors such as aggressiveness and altruism.

 \bibliographystyle{IEEEtran}
 \bibliography{ifacconf}   

 \end{document}